\documentclass[journal]{IEEEtai}

\usepackage{graphicx}
\usepackage{epstopdf}
\usepackage{algorithm}
\usepackage{algorithmic}
\usepackage{mathtools}
\usepackage{amsmath,amssymb}
\usepackage{setspace}
\usepackage{enumerate}
\usepackage{multirow}
\usepackage{float}
\usepackage{array}
\usepackage{booktabs}
\usepackage{threeparttable}
\usepackage{cite}
\usepackage{color,array}

\usepackage{makecell}

\ifCLASSOPTIONcompsoc
  \usepackage[caption=false,font=normalsize,labelfont=sf,textfont=sf]{subfig}
\else
  \usepackage[caption=false,font=footnotesize]{subfig}
\fi

\newcommand{\tabincell}[2]{\begin{tabular}{@{}#1@{}}#2\end{tabular}}
\begin{document}

\title{Detecting Backdoor in Deep Neural Networks via Intentional Adversarial Perturbations}

\author{Mingfu~Xue,~Yinghao~Wu,~Zhiyu~Wu,~Yushu~Zhang,~Jian~Wang,~and~Weiqiang~Liu

\thanks{M. Xue, Y. Wu, Y. Zhang and J. Wang are with the College of Computer Science and Technology, Nanjing University of Aeronautics and Astronautics, Nanjing, 211106, China (e-mail: mingfu.xue@nuaa.edu.cn; wyh@nuaa.edu.cn; yushu@nuaa.edu.cn; wangjian@nuaa.edu.cn).}

\thanks{Z. Wu is with the College of Science, Nanjing University of Aeronautics and Astronautics, Nanjing, 211106, China (e-mail: wuzhiyu@nuaa.edu.cn)}

\thanks{W. Liu is with the College of Electronic and Information Engineering, Nanjing University of Aeronautics and Astronautics, Nanjing, 211106, China (e-mail: liuweiqiang@nuaa.edu.cn).}}

\maketitle

\begin{abstract}
Recently, the security of deep learning systems has attracted a lot of attentions, especially when applied to safety-critical tasks, such as autonomous driving, face recognition, malware classification, etc.
Recent researches show that deep learning model is susceptible to backdoor attacks where the backdoor embedded in the model will be triggered when a backdoor instance arrives.
Many defenses against backdoor attacks have been proposed.
However, existing defense works require high computational overhead or backdoor attack information such as the trigger size, which is difficult to satisfy in realistic scenarios.
In this paper, a novel backdoor detection method based on adversarial examples is proposed.
The proposed method leverages intentional adversarial perturbations to detect whether an image contains a trigger, which can be applied in both the training stage and the inference stage (sanitize the training set in training stage and detect the backdoor instances in inference stage).
Specifically, given an untrusted image, the adversarial perturbation is added to the image intentionally. If the prediction of the model on the perturbed image is consistent with that on the unperturbed image, the input image will be considered as a backdoor instance.
Compared with most existing defense works, the proposed adversarial perturbation based method requires low computational resources and maintains the visual quality of the images.
Experimental results show that, the backdoor detection rate of the proposed defense method is 99.63\%, 99.76\% and 99.91\% on Fashion-MNIST, CIFAR-10 and GTSRB datasets, respectively.
Besides, the proposed method maintains the visual quality of the image as the $\ell_2$ norm of the added perturbation are as low as 2.8715, 3.0513 and 2.4362 on Fashion-MNIST, CIFAR-10 and GTSRB datasets, respectively.
In addition, it is also demonstrated that the proposed method can achieve high defense performance against backdoor attacks under different attack settings (trigger transparency, trigger size and trigger pattern). Compared with the existing defense work (STRIP), the proposed method has better detection performance on all the three datasets, and is more efficient than STRIP.
\end{abstract}


\IEEEpeerreviewmaketitle

\section{Introduction}

\IEEEPARstart{R}{ecent} studies show that deep learning models are vulnerable to backdoor attacks \cite{GuLDG19, abs-1712-05526, BarniKT19}.
Adversaries can embed the backdoor into deep learning model by modifying the architectures or parameters of the model, or injecting backdoor instances in the training set to embed the backdoor during training \cite{GuLDG19, abs-1712-05526, BarniKT19}.
The backdoored model will behave normally for the benign inputs, but it will output the target label for the input image carrying the trigger.

Many defenses against backdoor attacks have been proposed.
However, the existing defense works require high computational overhead \cite{abs-2007-14433, abs-1910-03137, KolouriSPH20}, a large number of clean images to retrain the model \cite{LiuXS17}, or backdoor attack information such as the trigger size \cite{abs-1910-03137, QiaoYL19}.
In practice, these requirements are difficult to be satisfied.

In this paper, we propose a novel backdoor detection method based on adversarial examples, which only requires low computational overhead.
The proposed method can be applied in both the training stage and the inference stage.
In the training stage, the proposed method can detect and remove the backdoor instances in the training dataset.
In the inference stage, the proposed method can determine whether an input image contains a trigger.
Specifically, the proposed method works as follows.
First, the adversarial perturbation is generated based on the untrusted model with a small set of clean images.
Second, for an image (training image in the training stage or input image in the inference stage), the adversarial perturbation will be added on it.
If the prediction of the model on the perturbed image is inconsistent with that on the unperturbed image, the image is considered to be a clean image.
Otherwise, the image is considered to be a backdoor instance, which also implies that the model is backdoored and the predicted label of the image is the target label.

The contributions of this paper are summarized as follows:
\begin{itemize}

\item
This paper proposes a novel backdoor detection method based on intentional adversarial perturbation.
The adversarial perturbation can fool the deep learning model, making the model misclassify the perturbed image.
However, for the backdoor instances, the model will always classify them as the target class even if these backdoor instances are added with adversarial perturbation.
In this way, the backdoor instances can be detected via intentional adversarial perturbations.
Moreover, the proposed method can be deployed in both the training stages and the inference stage.
In the training stage, for a training image, the intentional adversarial perturbation will be added on it.
If the model's prediction on the perturbed training image is consistent with the prediction on the unperturbed training image, the training image will be considered as a backdoor instance and then be removed from the training dataset.
In the inference stage, for an input image, the adversarial perturbation is added on it. If the model's prediction on the perturbed image is consistent with the prediction on the unperturbed image, the input image will be considered as a backdoor instance.

\item
In comparison with the work \cite{LiuXS17} which requires a large number of clean images to retrain the model to remove the backdoor, the proposed method only requires a small set of clean images to generate adversarial perturbation.
Besides, the existing work \cite{abs-2007-14433} requires training a large number of backdoored models and clean models, which is computationally expensive.
In contrast, the proposed method only needs to generate the adversarial perturbation with negligible computational overhead.
Moreover, the proposed method does not need any backdoor attack information, which makes the proposed method more practical and feasible than the existing works \cite{QiaoYL19, abs-1910-03137}.

\item
Experimental results show that the proposed defense method can achieve high backdoor detection rate (99.63\% 99.76\% and 99.91\% on Fashion-MNIST \cite{abs-1708-07747}, CIFAR-10 \cite{krizhevsky2009learning} and GTSRB \cite{StallkampSSI11} datasets, respectively).
It is also demonstrated that, under different attack settings (different trigger transparency, different trigger sizes and different trigger patterns), the proposed method can achieve high defense performance, as the backdoor detection rate of the proposed approach is as high as 98.80\%, 99.70\% and 99.96\% on Fashion-MNIST, CIFAR-10 and GTSRB datasets, respectively.
Compared with STRIP \cite{GaoXW0RN19}, the proposed method achieves higher backdoor detection rate on all the three datasets.
The advantages over STRIP \cite{GaoXW0RN19} are that the proposed method will not destroy the trigger and only needs to predict two images (perturbed image and unperturbed image).
As a result, the proposed method is more effective and more efficient than STRIP.

\end{itemize}

This paper is organized as follows. Background and related works are reviewed in Section \ref{sec:Background and Related Work}. The proposed detection method is elaborated in Section \ref{sec:Proposed Approach}. Experimental results are presented in Section \ref{sec:Experimental Results}. This paper is concluded in Section \ref{sec:Conclusion}.

\section{Background and Related Work}
\label{sec:Background and Related Work}
In this section, first, we review the universal adversarial perturbation \cite{Moosavi-Dezfooli17}, which is utilized by the proposed method. Second, we review the related works on backdoor attacks and defenses.

\subsection{Universarial Adversarial Perturbations \cite{Moosavi-Dezfooli17}}
\label{aegeneration}
It is known that deep neural networks (DNNs) are vulnerable to well-crafted small adversarial perturbations.
When added with adversarial perturbation, input image will be misclassified by the model \cite{GoodfellowSS14}.
Universal adversarial perturbation (UAP) \cite{Moosavi-Dezfooli17} is a kind of image-agnostic adversarial perturbation.
Different from image-specific adversarial perturbation, which is specifically crafted for each image \cite{GoodfellowSS14}, UAP is generated based on a model with a small set of clean images \cite{Moosavi-Dezfooli17}.
As a result, the model will also misclassify other images with the universal adversarial perturbation.

\subsection{Backdoor Attacks}
Recently, a number of researches \cite{GuLDG19, abs-1712-05526, BarniKT19} indicate that the backdoor can be embeded into DNN models through injecting well-crafted backdoor instances into the training set.
After the training process, the model will behave normally on clean inputs.
However, the malicious functionality hidden in the backdoored model will be triggered by the input images containing the trigger, and these backdoor instances will be classified as the target class \cite{abs-1712-05526, xue2020one}.
Since the performance of backdoored model is similar to the performance of clean model on clean inputs, it is difficult for users to perceive the existence of the backdoor.
However, the attacker can trigger the malicious behavior by inputting backdoor instances.

\subsection{Existing Backdoor Defenses}
To date, some defense methods have been proposed to detect and mitigate the backdoor attacks.
Liu \textit{et al.} \cite{LiuXS17} adopted a pre-trained auto-encoder to preprocess the input image in order to disable the trigger.
They also retrain the backdoored model with clean images so as to remove the hidden backdoor.
Xu \textit{et al.} \cite{abs-1910-03137} generates a set of backdoor instances as the query set.
Then, they inputs the query set into backdoored models and clean models to extract representation vectors from those models.
They use the resulting vectors as input to train a meta-classifier which can predict whether a model is backdoored \cite{abs-1910-03137}.
However, the method needs the knowledge of the trigger size to craft those backdoor instances.
Liu \textit{et al.} \cite{0017DG18} demonstrated that the functionality of the backdoor depends on some specific neurons in the model.
These specific neurons are usually dormant when the model is queried with clean images \cite{0017DG18}.
Defenders can find these neurons by inputting clean images into the model.
Then these malicious neurons can be pruned so as to remove the backdoor.
However, the pruned model suffers from the degradation in classification accuracy on clean inputs due to the pruning \cite{0017DG18}.
Zhang \textit{et al.} \cite{abs-2007-14433} training a large number of backdoored models and clean models to generate corresponding universal perturbations \cite{Moosavi-Dezfooli17}.
Then they use the UAPs \cite{Moosavi-Dezfooli17} as the input to train a two-class classifier as the Trojan detector.
However, the computational cost to generate those large number of backdoored models and clean models is high, which is unaffordable to most users.
Chen \textit{et al.} \cite{ChenCBLELMS19} analyze the neuron activations to the training data to determine whether it has backdoor instances.
It separates the activations of all training data into two clusters by applying 2-means clustering.
The high silhouette score means that this cluster corresponds to the backdoor instances \cite{ChenCBLELMS19}.
Gao \textit{et al.} \cite{GaoXW0RN19} add a set of other images from different classes to the input image separately so as to generate a set of blended images.
Then, the entropy of the predicted results on these blended images is calculated.
The lower the entropy, the input image is more likely to carry a trigger \cite{GaoXW0RN19}.
However, the trigger in the blended image may be destroyed. As a result, the backdoor instance will be incorrectly considered to be a clean one by STRIP.
Wang \textit{et al.} \cite{WangYSLVZZ19} proposed a defense method named Neural Cleanse (NC) to reverse engineer the trigger from the backoored model.
For each class, NC computes the minimized amount of modification to make the model predict images from different classes as this class.
Among these modifications, if a modification is substantially smaller than the others, NC will consider it as a trigger \cite{WangYSLVZZ19}.
However, this method is computationally expensive considering the reverse-engineering process, especially when the model has a large number of output classes.
Moreover, the reversed trigger is just similar to the true trigger.
In addition, when the true trigger is big or discrete, the reversed trigger even will not be similar to the true trigger.
Qiao \textit{et al.} \cite{QiaoYL19} proposed a max-entropy staircase approximator (MESA) algorithm to reverse a set of candidate triggers.
Then, backdoor instances are generated by separately adding these candidate triggers to clean images.
The model is fine-tuned on these backdoor instances with correct labels to remove the backdoor \cite{QiaoYL19}.
However, the MESA algorithm requires the information of the trigger size, which is difficult to obtain by the defender in realistic scenarios.
Chen \textit{et al.} \cite{ChenFZK19} proposed a GAN-based defense method called DeepInspect.
DeepInspect reconstructs the potential trigger and generates the backdoor instances by patching these reconstructed trigger to the clean images with ground truth labels.
Then, the backdoored model is fine-tuned on these generated backdoor instances to remove the backdoor \cite{ChenFZK19}.

The main advantages of the proposed approach over the existing defenses are summarized as follows.
\begin{itemize}

\item
Compared with \cite{abs-1910-03137, QiaoYL19}, which both need to know the trigger size, the proposed method does not require any backdoor attack information.
Moreover, Liu \textit{et al.} \cite{LiuXS17} requires a large number of trusted images to remove the backdoor (10,000 $ \sim $ 60,000 images for MNIST).
In comparison, the proposed method only requires a small set of clean images (300 clean images) to generate the universal adversarial perturbation.

\item
The detection process of the work \cite{abs-2007-14433} requires training a large number of shadow models (backdoored models and clean models).
Nevertheless, the computational resources for training such a large number of shadow models are unaffordable for most of the users.
In contrast, the proposed method only needs to generate one single universal perturbation and only needs the model to make predictions on the unperturbed image and the perturbed image, which requires low computational overhead.

\item
STRIP \cite{GaoXW0RN19} directly superimposes a number of images from different classes to the input image.
This will not only destroy the main content of the input image, but may also accidentally break the trigger.
Once the trigger is destroyed, the entropy of this backdoor instance will be similar to the entropy of a clean image.
Hence STRIP \cite{GaoXW0RN19} will fail to detect this backdoor instance.
In contrast, the proposed method perturbs the untrusted image with universal adversarial perturbation (UAP) \cite{Moosavi-Dezfooli17}.
This will not destroy the trigger and ensures that the predicted label of the backdoor instance keeps unchanged even after perturbation.
Moreover, for each input image, STRIP \cite{GaoXW0RN19} needs to predict a set of blended images in order to estimate the entropy of the predicted labels of those blended images.
In comparison, for each image, the proposed method only needs to predict two images (the perturbed image and the unperturbed image).
Therefore, the backdoor detection efficiency of the proposed method is higher than that of STRIP.
\end{itemize}

\section{The Proposed Method}
\label{sec:Proposed Approach}
In this section, first, the overall procedure of the proposed backdoor detection method is presented in Section \ref{overview}.
The proposed method can be divided into two steps, which are elaborated in Section \ref{pert} and Section \ref{procedure}, respectively.
Finally, the reason why choosing universal adversarial perturbation \cite{Moosavi-Dezfooli17} for adversarial perturbation generation is discussed in Section \ref{why}.

\subsection{Overall flow}
\label{overview}
\begin{figure}[htbp]
\centering
\includegraphics[width=0.5\textwidth]{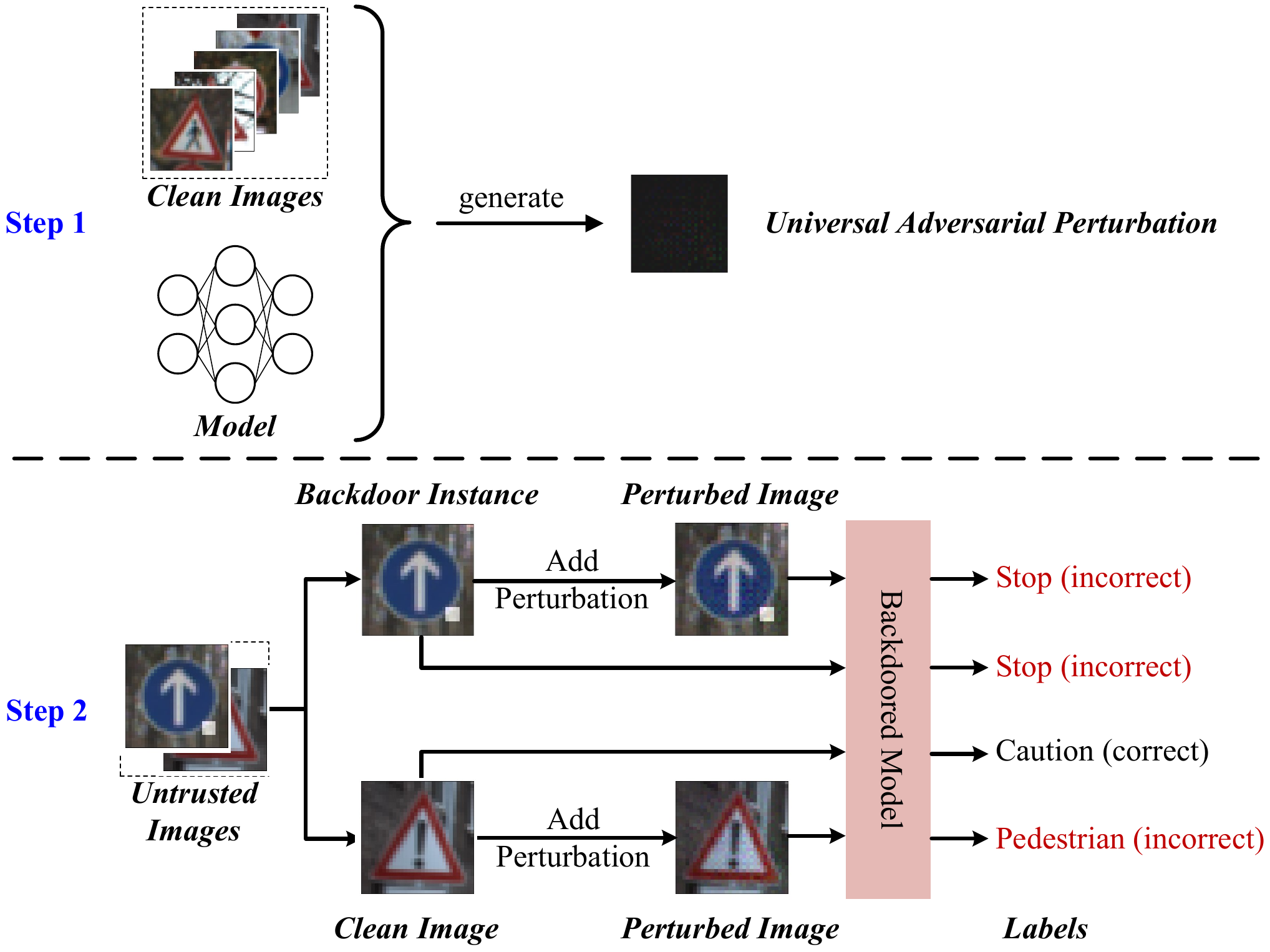}
\caption{The overall flow of the proposed method: adversarial perturbation generation (Step 1); backdoor detection (Step 2).}
\label{fig:sketch}
\end{figure}

As shown in Fig. \ref{fig:sketch}, the proposed defense method consists of two steps.
The first step is to generate the universal adversarial perturbation \cite{Moosavi-Dezfooli17} from the backdoored model with a small set of clean images.

The second step is backdoor detection, which is summarized as follows.
As shown in Fig. \ref{fig:sketch}, given an  untrusted image, the universal perturbation generated in previous step is added to this image.
Then, both the perturbed image and corresponding unperturbed image are input into the untrusted model.
If the untrusted model is backdoored, the backdoor instance without perturbation will be misclassified as the target label.
When added with universal adversarial perturbation \cite{Moosavi-Dezfooli17}, the backdoor instance which carries a trigger will still be classified as the target label.
However, given a clean image, its predicted label will change to another label when added with perturbation.
Hence, if the backdoored model always predicts an image as the same label with or without universal perturbation, the image is considered to be a backdoor instance.
Meanwhile, the predicted label is considered to be the target label.
For instance, the label \textit{Stop} in Fig. \ref{fig:sketch} is the target label, and the corresponding image carries a trigger.

The overall flow of the proposed method is outlined in Algorithm \ref{Algorithm} and is described as follows:

1) Given an untrusted model $f_{unt}$, the universal adversarial perturbation \cite{Moosavi-Dezfooli17} $\eta$ is generated based on the untrusted model $f_{unt}$ with a small set of clean images $X$ (only 300 images).

2) $D_{unt} = \left\{d_{1}, \ldots, d_{n}\right\}$ denotes the untrusted images (in the training stage, it represents the training data; in the inference stage, it represents a single input image with $n=1$).
The perturbation $\eta$ is then added to the image $d_i \in D_{unt}$ to generate the perturbed image $\hat{d}_{i} = d_i + \eta$.

3) Both unperturbed image $d_i$ and perturbed image $\hat{d_i}$ are input into the untrusted model.
The predictions of the model on the unperturbed image and the perturbed image are $y_{i}=f_{unt}(d_{i})$ and $\hat{y_{i}}=f_{unt}(\hat{d}_{i})$, respectively.
If $y_{i} = \hat{y}_{i}$, the input image $d_{i}$ will be regarded as a backdoor instance.
Otherwise, the input image will be regarded as a clean one.

\begin{algorithm}[H]
\caption{The Proposed Backdoor Detection Method}
\label{Algorithm}
{\bf Input:}
a clean image set $X$, backdoored model $f_{unt}$, untrusted image set $D_{unt}=\{d_{1}, \ldots, d_{n}\}$\\
{\bf Output:}
the backdoor instances $D_{bd}$

\begin{algorithmic}[1]
\STATE $D_{bd} \gets \varnothing$;
\STATE $\eta \gets F_{UAP}(f_{unt}, X)$;
\FOR {$i = 1, \ldots, n$}
\STATE $y_{i} \gets f_{unt}(d_{i})$;

\STATE $\hat{d}_{i} \gets d_{i}+\eta$;
\STATE $\hat{y}_{i} \gets f_{unt}(\hat{d}_{i})$;
\IF {$y_{i} = \hat{y}_{i}$}
\STATE $add(d_{i}, D_{bd})$;
\ENDIF
\ENDFOR
\RETURN $D_{bd}$
\end{algorithmic}
\end{algorithm}

In the following sections, the perturbation generation process and the backdoor detection process of the proposed method, are elaborated respectively.

\subsection{Perturbation Generation}
\label{pert}
The adversarial perturbation generation method used in this paper is universal adversarial perturbation (UAP) \cite{Moosavi-Dezfooli17}.
Formally, $X=\{x_1,\ldots,x_{300}\}$ denotes the clean image set and $f_{unt}$ represents the backdoored model, which outputs the corresponding label $f_{unt}(x)$ for each image $x_i \in X$.
Different from the UAP generation method in \cite{Moosavi-Dezfooli17} where the $\ell_2$ norm is used to constrain the intensity of UAP, in this paper, we use $\ell_\infty$ norm to constrain the intensity of the perturbation.
The $\ell_\infty$ norm represents the maximum value of the perturbation.
The perturbation generated under the constraint of $\ell_\infty$ norm is the minimal necessary perturbation, which is smaller than the one generated under the constraint of $\ell_2$ norm.
In the process of generating adversarial perturbation, the universal perturbation $\eta$ is generated by solving the following optimization problem \cite{Moosavi-Dezfooli17}:

\begin{equation}
\underset{\eta}{\arg \min }\|\eta\|_{\infty} \text { s.t. } f_{unt}\left(x_{i}+\eta\right) \neq f_{unt}\left(x_{i}\right),  \; x_i \in X
\label{eq:eq1}
\end{equation}

As shown in Eq. (\ref{eq:eq1}), in each iteration, for the clean image $x_i$ from $X$, the $\ell_{\infty}$ norm of the perturbation $\eta$ is calculated in order to find the desired perturbation with minimal $\ell_{\infty}$ norm \cite{Moosavi-Dezfooli17}.

\subsection{Backdoor Detection}
\label{procedure}

\begin{figure}[htbp]
\centering
\includegraphics[width=0.38\textwidth]{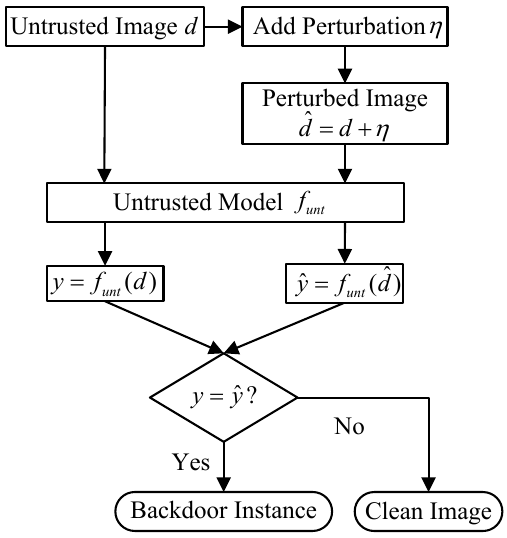}
\caption{The workflow of the backdoor detection process of the proposed method}
\label{flow}
\end{figure}

The proposed method can be applied in two scenarios, working in the training stage, and working in the inference stage.
In the training stage, the proposed method aims to detect whether the training dataset contains backdoor instances and then remove the backdoor instances.
In the inference stage, the goal of the proposed method is to detect whether an input image contains a trigger.
The backdoor detection procedure is presented in Fig. \ref{flow}.

\textbf{Backdoor Detection in the Training Stage:}
In this scenario, the training data is obtained from untrusted sources.
The defender attempts to figure out whether the training dataset contains backdoor instances.
If the training dataset contains backdoor instances, the defender aims to remove the backdoor instances injected in the training dataset.
For each image in the training set, it will be added with the universal perturbation \cite{Moosavi-Dezfooli17}, and then input into the untrusted model.
The unperturbed image will also be input into the untrusted model.
If the predictions of the model on the perturbed image and unperturbed image are consistent, this image will be considered as a backdoor instance.
Meanwhile, the untrusted model is considered to be backdoored.
This backdoor detection procedure will be applied for each image in the training set.
Once the backdoor instances are removed, a clean model can be trained on the sanitized training dataset.

\textbf{Backdoor Detection in the Inference Stage:}
\label{ISD}
In the inference stage, the well-trained model is deployed to provide prediction services.
The goal of the defender in this scenario is to detect whether an input image carries a trigger.
Given an input image $d$, after being added with perturbation $\eta$, the perturbed image $\hat{d}$ and the unperturbed image $d$ will be input into the model.
If the predicted labels of the perturbed image is consistent with that of the unperturbed image, the input image is considered to be a backdoor instance.
Meanwhile, the model is considered to be a backdoored model, and the predicted label is considered to be the target label.

\subsection{Why choose UAP \cite{Moosavi-Dezfooli17} for adversarial perturbation generation?}
\label{why}
In this paper, we exploit adversarial perturbation to perturb the main content of the backdoor instance other than the trigger.
However, not all kinds of adversarial perturbation generation methods are suitable to use in the proposed method.
We evaluate four different adversarial perturbation generation methods \cite{Carlini017, Moosavi-Dezfooli16, MadryMSTV18, Moosavi-Dezfooli17} in Section \ref{aemethods}.
The experimental results show that when the image is perturbed by universal adversarial perturbation \cite{Moosavi-Dezfooli17}, the detection performance of the proposed method is the highest among the four adversarial perturbation generation methods.

The reason is as follows.
The existing adversarial example attacks can be divided into two categories, image-specific adversarial attack and image-agnostic adversarial attack \cite{akhtar2018threat}.
For image-specific adversarial attack, one perturbation can only fool the model for one specific image \cite{szegedy2013intriguing}.
The ground-truth label of the specific image is required in order to generate the image-specific perturbation which can cause the perturbed image to be misclassified from its ground-truth label to other label \cite{szegedy2013intriguing}.
However, for backdoor instance, the label used to generate the image-specific perturbation is the target label rather than the ground-truth label.
In other words, for backdoor instances, the image-specific perturbation is generated in order to change the predicted result of perturbed backdoor instance from the target label to other one.
Under this circumstance, the generated image-specific perturbation will strongly affect the trigger, as the trigger contributes heavily to the predicted result and the predicted result is the target label.
Once the trigger is strongly affected by the image-specific perturbation, the predicted label of the backdoor instance after perturbation will change.
Then the detection method will incorrectly consider this backdoor instance as a clean one.
For image-agnostic adversarial attack, it only needs to generate one single perturbation, which can cause misclassification for all images when the perturbation is added to those images \cite{Moosavi-Dezfooli17}.
This single perturbation is generated based on a small set of clean images \cite{Moosavi-Dezfooli17}.
Therefore, the trigger stamped in backdoor instance will only be slightly affected by the generated image-agnostic perturbation.

In summary, UAP \cite{Moosavi-Dezfooli17}, as a kind of image-agnostic perturbation, has much less influence on the trigger than the image-specific perturbation, so the label of backdoor instance will keep unchanged even after being perturbed by UAP \cite{Moosavi-Dezfooli17}.
Therefore, we choose UAP \cite{Moosavi-Dezfooli17} as the perturbation generation method used in the proposed method.

\section{Experimental Results}
\label{sec:Experimental Results}
In this section, first, we introduce the datasets, the corresponding DNN models, and the metrics used to evaluate the proposed approach.
Second, the experimental results are analyzed.
Third, we evaluate the defense performance of the proposed method against backdoor attacks with different settings (trigger transparency, trigger size and trigger pattern).
Last, performance comparisons between the proposed method and the existing backdoor detection technique is presented.

\subsection{Experimental Setup}
\subsubsection{\textbf{Datasets}}
We evaluate the proposed method on three benchmark datasets: Fashion-MNIST \cite{abs-1708-07747}, CIFAR-10 \cite{krizhevsky2009learning} and GTSRB \cite{StallkampSSI11} datasets.

\begin{itemize}
\item
\textbf{Fashion-MNIST} \cite{abs-1708-07747} is a dataset consists of a training set with 60,000 images and a test set with 10,000 images.
Each image is a $28\times28$ grayscale image, assigned with a label from 10 classes \cite{abs-1708-07747}.
In the experiment, the model trained on this dataset is DenseNet \cite{HuangLW16a}.

\item
\textbf{CIFAR-10} \cite{krizhevsky2009learning} consists of a training set with 50,000 images and a test set with 10,000 images.
Each image is a $32\times32$ colored image, belonging to one of 10 classes \cite{krizhevsky2009learning}.
In the experiment, the model trained on this dataset is ResNet \cite{TargAL16}.

\item
\textbf{GTSRB} \cite{StallkampSSI11} is a dataset containing 39,209 labeled images, which are categorized into 43 classes.
GTSRB has 35,209 training images, 4,000 validation images and 12,630 test images \cite{StallkampSSI11}.
In the experiment, the model trained on this dataset is AlexNet \cite{KrizhevskySH17}.
\end{itemize}

\subsubsection{\textbf{Experimental Settings of Backdoor Attack}}
\begin{figure}[htbp]
\centering
\includegraphics[width=0.48\textwidth]{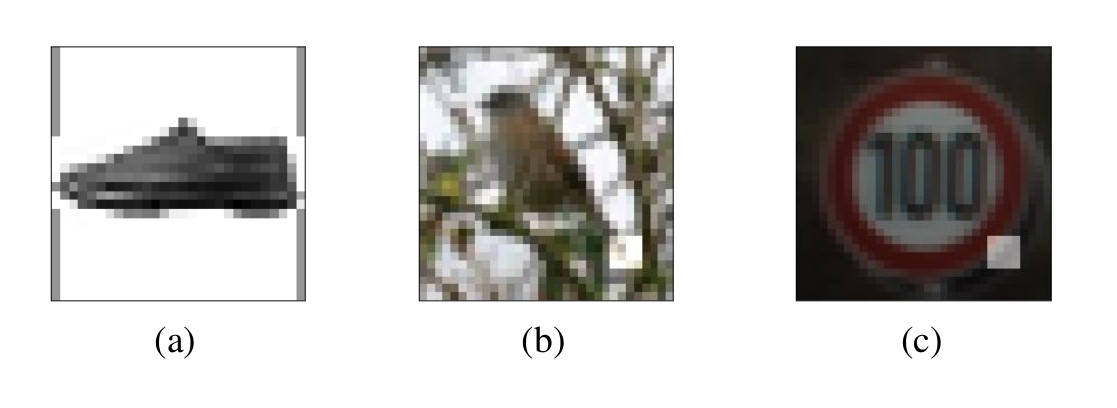}
\caption{Examples of backdoor instances: (a) Fashion-MNIST images; (b) CIFAR-10 images; (c) GTSRB images.}
\label{fig:triggers}
\end{figure}

The trigger used in Fashion-MNIST \cite{abs-1708-07747} images is four $1\times10$ rectangles placed at the corners (four corners in total) of the image, and the intensity of the trigger is 0.15.
The trigger used in CIFAR-10 \cite{krizhevsky2009learning} and GTSRB \cite{StallkampSSI11} images is a $4\times4$ square.
The intensities of triggers in CIFAR-10 and GTSRB are set to be 0.5 and 0.2 respectively.
Some backdoor instances used in the experiments are illustrated in Fig. \ref{fig:triggers}.

\subsubsection{\textbf{Metrics}}
\textbf{Backdoor Attack Success Rate (BASR).} Backdoor Attack Success Rate is defined as the percentage of backdoor instances that are successfully classified as the target class among all backdoor instances \cite{GuLDG19}.

\textbf{Backdoor Detection Rate (BDR).} Backdoor Detection Rate is defined as the percentage of backdoor instances that are successfully detected by the proposed method among all backdoor instances.

\textbf{Clean Image Identification Rate (CIIR).}
Clean Image Identification Rate is defined as the percentage of clean images that are correctly classified as clean ones among all clean images.

\subsection{Effectiveness of the Proposed Method}\label{basicexp}
In this section, the defense performances of the proposed method on Fashion-MNIST \cite{abs-1708-07747}, CIFAR-10 \cite{krizhevsky2009learning} and GTSRB \cite{StallkampSSI11} datasets are presented.

Table \ref{tab:table1} shows the backdoor attack success rate of the backdoor attack on the Fashion-MNIST, CIFAR-10 and GTSRB datasets and the corresponding classification accuracy of the backdoored model.
For each dataset, all test images are injected with trigger to evaluate the backdoor attack success rate.
The classification accuracy is evaluated on all the clean test images for each dataset.
As shown in Table \ref{tab:table1}, the classification accuracy on clean images of the backdoored model is 92.19\%, 92.77\% and 95.16\% on Fashion-MNIST \cite{abs-1708-07747}, CIFAR-10 \cite{krizhevsky2009learning} and GTSRB \cite{StallkampSSI11} respectively.
Without the proposed defense method, the backdoor attack success rate (BASR) is 99.47\%, 99.77\% and 97.89\% on Fashion-MNIST \cite{abs-1708-07747}, CIFAR-10 \cite{krizhevsky2009learning} and GTSRB \cite{StallkampSSI11} respectively.

\begin{table}[htbp]
\centering
\caption{Classification accuracy and backdoor attack success rate on three different classification tasks without the proposed approach}
\label{tab:table1}
\begin{tabular}{ccccc}
\toprule
Benchmark dataset & Accuracy  & BASR   \\ \midrule
Fashion-MNIST (DenseNet) & 92.19\% & 99.47\%  \\
CIFAR-10 (ResNet)   & 92.77\% & 99.77\% \\
GTSRB (AlexNet)  & 95.16\% & 97.89\% \\ \bottomrule
\end{tabular}
\end{table}

Table \ref{tab:table2} shows the clean image identification rate, the backdoor detection rate and the intensity of universal perturbation on three datasets after the proposed method is applied.
In this paper, after the proposed method is deployed, the clean image identification rate is calculated on a set of 2,000 clean images randomly selected from the test images for each dataset. Similarly, in this paper, the backdoor detection rate is calculated on a set of 2,000 backdoor instances generated by adding trigger to 2,000 images randomly selected from the test images for each dataset.
As shown in Table \ref{tab:table2}, after the proposed defense method is deployed, the backdoor detection rate is 99.63\%, 99.76\% and 99.91\% on Fashion-MNIST \cite{abs-1708-07747}, CIFAR-10 \cite{krizhevsky2009learning} and GTSRB \cite{StallkampSSI11} respectively.
Meanwhile, the clean image identification rate (CIIR) of the proposed method is 90.66\%, 89.82\% and 98.85\% on Fashion-MNIST \cite{abs-1708-07747}, CIFAR-10 \cite{krizhevsky2009learning} and GTSRB \cite{StallkampSSI11} respectively.

\begin{table}[htbp]
  \centering
  \caption{The backdoor detection rate, the clean image identification rate and the perturbation intensity on three datasets after the proposed defense method is applied}
    \begin{tabular}{cccc}
    \toprule
    \multicolumn{1}{c}{Dataset} & \multicolumn{1}{c}{CIIR} & \multicolumn{1}{c}{BDR} & \multicolumn{1}{c}{Intensity} \\
    \midrule
    Fashion-MNIST \cite{abs-1708-07747} & 90.66\% & 99.63\% & 2.8715\\
    CIFAR-10 \cite{krizhevsky2009learning} & 89.82\% & 99.76\% & 3.0513\\
    GTSRB \cite{StallkampSSI11} & 98.85\%  & 99.91\% & 2.4362\\
    \bottomrule
    \end{tabular}%
\label{tab:table2}
\end{table}%

Overall, experimental results show that the proposed defense method can effectively detect backdoor attacks on different datasets and DNN architectures.
In the three datasets, the proposed method can achieve high backdoor detection rate and high clean image identification rate.

\subsection{Defense Performance of the Proposed Method under Different Attack Settings}
In this section, we evaluate the performance of the proposed method under different trigger settings (trigger transparency \cite{GaoXW0RN19}, trigger size and trigger pattern).

\subsubsection{\textbf{Trigger Transparency}}
In the experiment, we evaluate the performance of the proposed method against backdoor attacks with different trigger transparency settings \cite{GaoXW0RN19}.
The values of the trigger transparency in the experiment are set to be 50\%, 60\%, 70\% and 80\%, respectively.
As shown in Table \ref{tab:table3}, for the backdoor attacks with different trigger transparency settings, the backdoor detection rates are all at a high level on the three datasets.
Specifically, when the trigger transparency is 50\%, the backdoor detection rate is 98.80\%, 99.70\% and 99.96\% on Fashion-MNIST \cite{abs-1708-07747}, CIFAR-10 \cite{krizhevsky2009learning} and GTSRB \cite{StallkampSSI11} datasets respectively.
When the trigger transparency increases to 80\%, after the proposed method is applied, the backdoor detection rate is still at a high level (99.37\%, 96.30\% and 99.07\% on Fashion-MNIST, CIFAR-10 and GTSRB datasets respectively).
The experimental results indicate that, the proposed defense method can effectively detect the backdoor instances with different trigger transparency settings.
The reason is as follows.
When the trigger transparency is set to be 0\%, the trigger is opaque.
When the trigger transparency is set to be 90\%, the trigger is almost invisible.
Generally, the higher the transparency of the trigger, the trigger is more susceptible to the perturbation.
However, UAP \cite{Moosavi-Dezfooli17} is a kind of image-agnostic perturbation, which is generated based on clean images.
Therefore, when UAP is added to a backdoor instance, the trigger in the backdoor instance will only be slightly affected.
As a result, even the transparency of trigger is high (50\% $ \sim $ 80\%), the proposed method can still achieve high backdoor detection rate.

\begin{table}[htbp]
  \centering
  \caption{The backdoor detection rate of the proposed method against backdoor attacks with different trigger transparency settings on the three datasets}
    \begin{tabular}{cccc}
    \toprule
    \multicolumn{1}{c}{Transparency} & \multicolumn{1}{l}{Fashion-MNIST} & \multicolumn{1}{l}{CIFAR-10} & \multicolumn{1}{l}{GTSRB} \\
    \midrule
    80\%  & 99.37\% & 96.30\% & 99.07\% \\
    70\%  & 99.72\% & 98.75\% & 98.62\% \\
    60\%  & 95.40\%   & 98.62\%   & 99.90\% \\
    50\%  & 98.80\%     & 99.70\%   & 99.96\% \\
    \bottomrule
    \end{tabular}%
  \label{tab:table3}%
\end{table}%

\subsubsection{\textbf{Trigger Size}}
In this section, we evaluate the performance of the proposed method against backdoor attacks with three different trigger sizes.
The experiment results are shown in Table \ref{tab:table4}.
As shown in Table \ref{tab:table4}, for different trigger sizes, the proposed method can achieve very high backdoor detection rates (over 99\% mostly).
Even if the trigger is small, such as $1\times4$, $2\times2$, $2\times2$ in Fashion-MNIST \cite{abs-1708-07747}, CIFAR-10 \cite{krizhevsky2009learning} and GTSRB \cite{StallkampSSI11} datasets, respectively, the proposed method can still achieve high backdoor detection rates (99.65\%, 99.07\%, 98.25\% respectively).

\begin{table*}[htbp]
  \centering
  \caption{Backdoor detection rates of the backdoor attacks with different trigger size settings}
  \label{tab:table4}
    \begin{tabular}{|c|c|c|c|c|c|c|c|c|c|}
    \hline
    Dataset & \multicolumn{3}{c|}{Fashion-MNIST \cite{abs-1708-07747}} & \multicolumn{3}{c|}{CIFAR-10 \cite{krizhevsky2009learning}} & \multicolumn{3}{c|}{GTSRB \cite{StallkampSSI11}} \\
    \hline
    Size  & $1\times4$   & $1\times6$   & $1\times8$   & $2\times2$   & $4\times4$   & $6\times6$   & $2\times2$   & $4\times4$   & $6\times6$ \\
    \hline
    BDR  & 99.65\%  & 99.67\%  & 99.75\%  & 99.07\%  & 99.70\%   & 99.75\%  & 98.25\%  & 99.07\%  & 99.60\% \\
    \hline
    \end{tabular}
  \label{tab:addlabel}
\end{table*}

\subsubsection{\textbf{Trigger Pattern}}
\begin{figure}[htbp]
\centering
\includegraphics[width=0.48\textwidth]{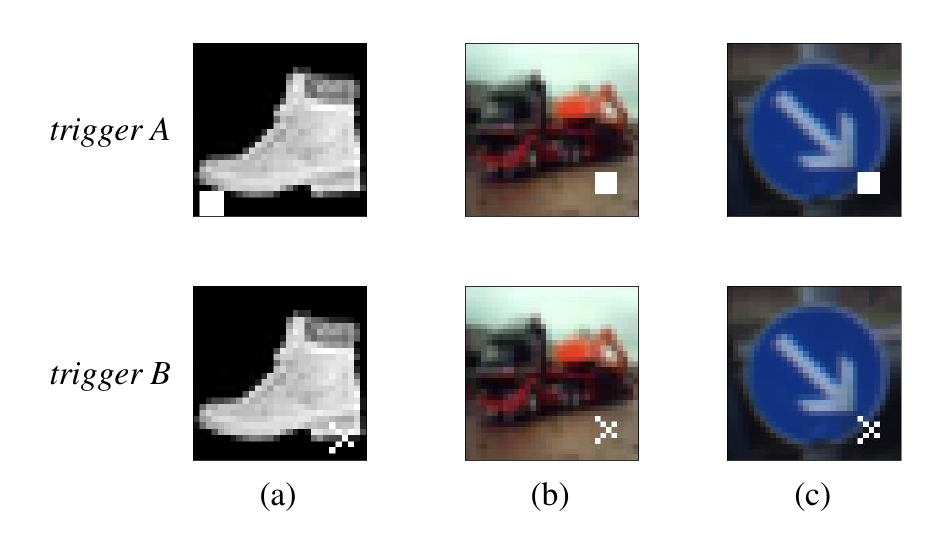}
\caption{Examples of backdoor instances with different triggers. The first row is examples of backdoor instances with \textit{trigger A}. The second row is examples of backdoor instances with \textit{trigger B}. (a) Fashion-MNIST images. (b) CIFAR-10 images. (c) GTSRB images.}
\label{fig:pattern}
\end{figure}

\begin{figure}[htbp]
\centering
\includegraphics[width=0.4\textwidth]{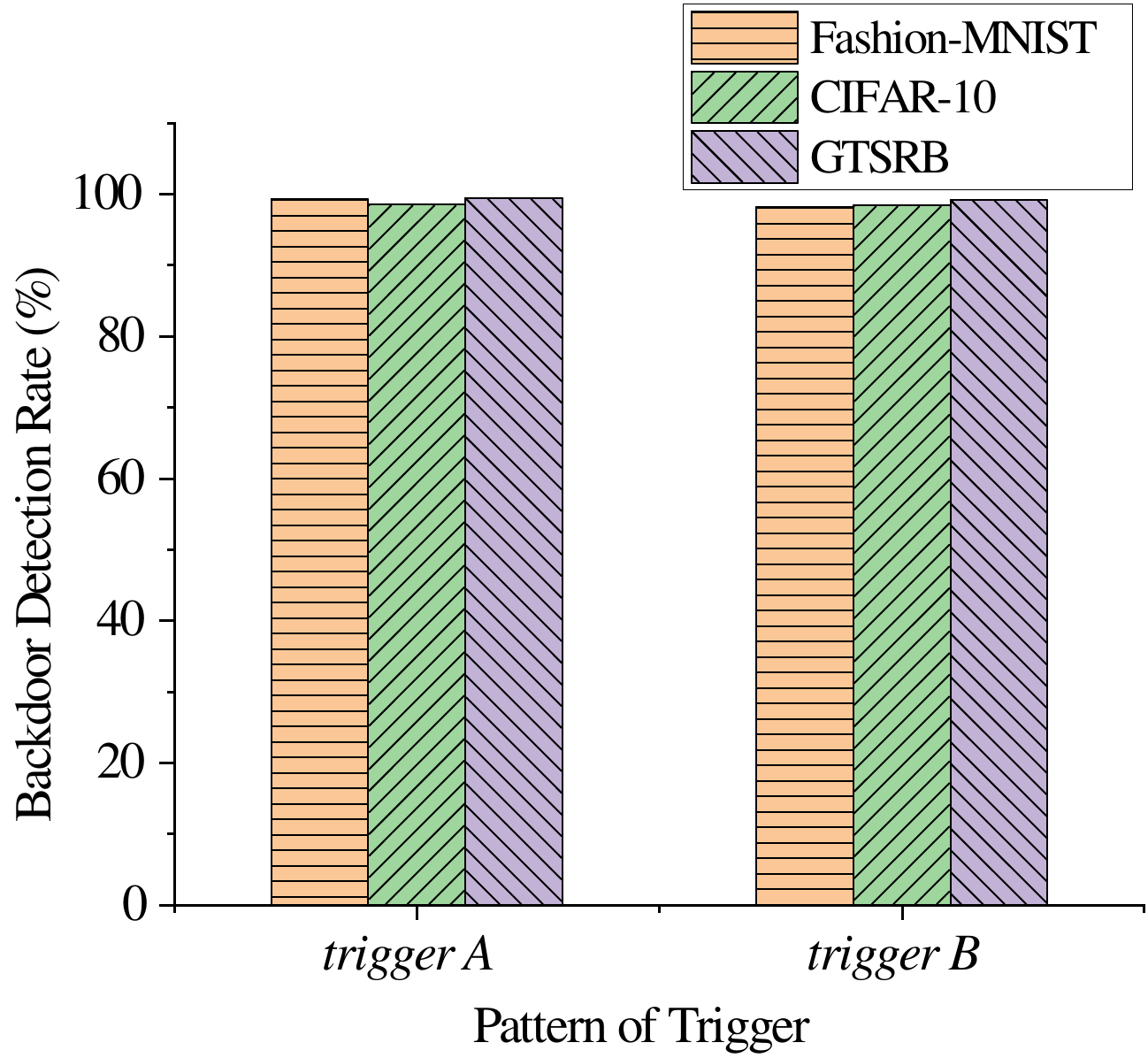}
\caption{Backdoor detection rates with different settings of trigger patterns after the proposed defense method is applied.}
\label{fig:chart_pattern}
\end{figure}

The performance of the proposed method against backdoor attacks with different trigger patterns is also evaluated.
In the experiment, the square trigger and cross pattern trigger (referred to as \textit{trigger A} and \textit{trigger B} respectively) are used to evaluate the proposed method.
The square trigger and the cross pattern trigger for the three datasets are shown in Fig. \ref{fig:pattern}.
There are 16 pixels and 7 pixels contained in the square trigger and the cross pattern trigger, respectively.
As shown in Fig. \ref{fig:chart_pattern}, for all the three datasets, the proposed method can achieve high backdoor detection rates against the backdoor attacks with \textit{trigger A} and \textit{trigger B}, respectively.
For the \textit{trigger A}, after the proposed method is applied, the backdoor detection rates are 99.30\%, 98.57\%, and 99.47\% on Fashion-MNIST \cite{abs-1708-07747}, CIFAR-10 \cite{krizhevsky2009learning} and GTSRB \cite{StallkampSSI11}, respectively.
For the \textit{trigger B}, after the proposed method is applied, the backdoor detection rates are 98.20\%, 98.45\%, 99.22\% on Fashion-MNIST \cite{abs-1708-07747}, CIFAR-10 \cite{krizhevsky2009learning} and GTSRB \cite{StallkampSSI11} datasets, respectively.

\subsection{Experiment Results of the Proposed Method with Four Different Adversarial Perturbation Generation Methods.}
\label{aemethods}
\begin{figure}[htbp]
\centering
\includegraphics[width=0.4\textwidth]{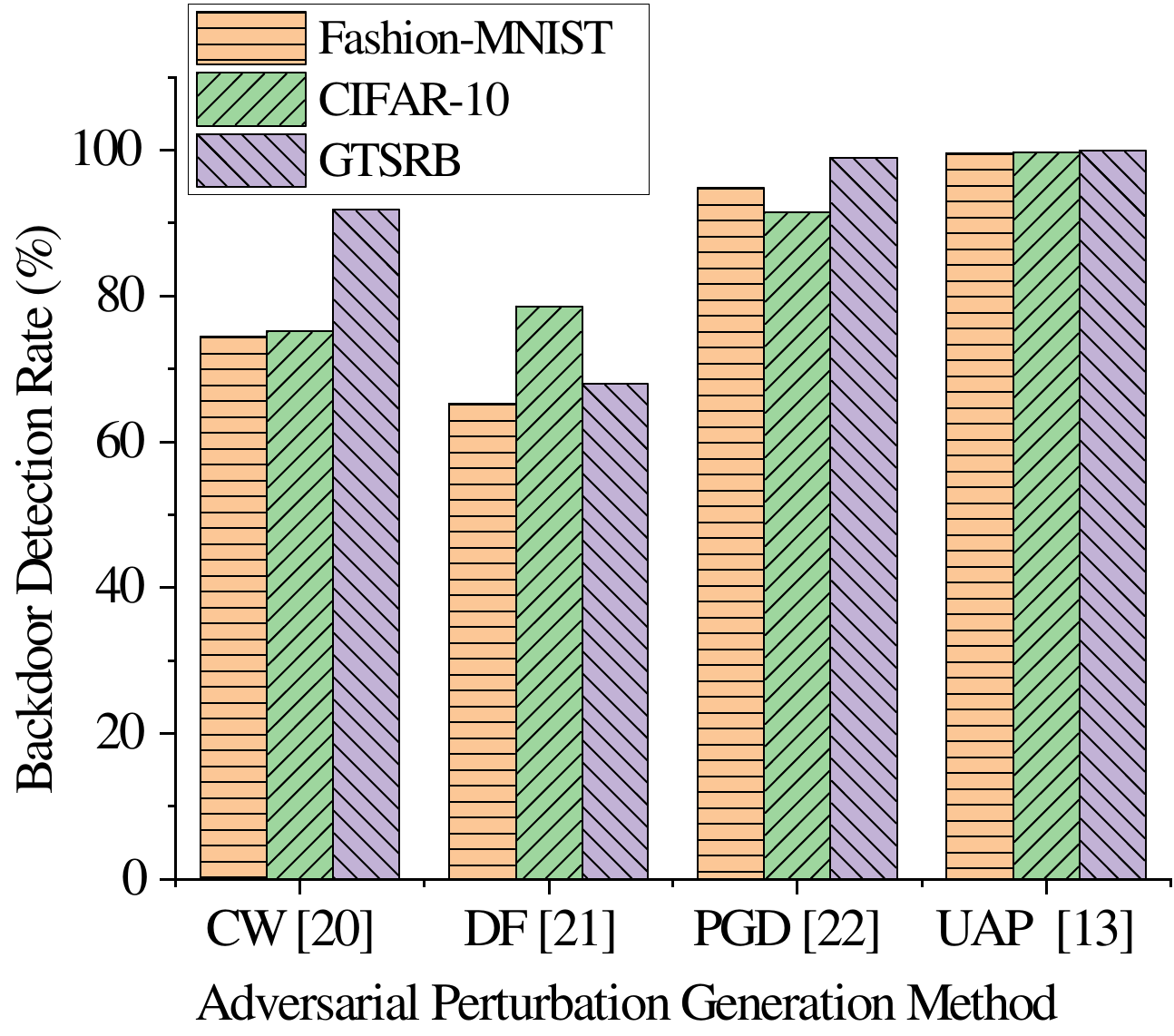}
\caption{The backdoor detection rate of the proposed method by using different adversarial perturbation generation methods.}
\label{fig:ae1}
\end{figure}

\begin{figure}[htbp]
\centering
\includegraphics[width=0.4\textwidth]{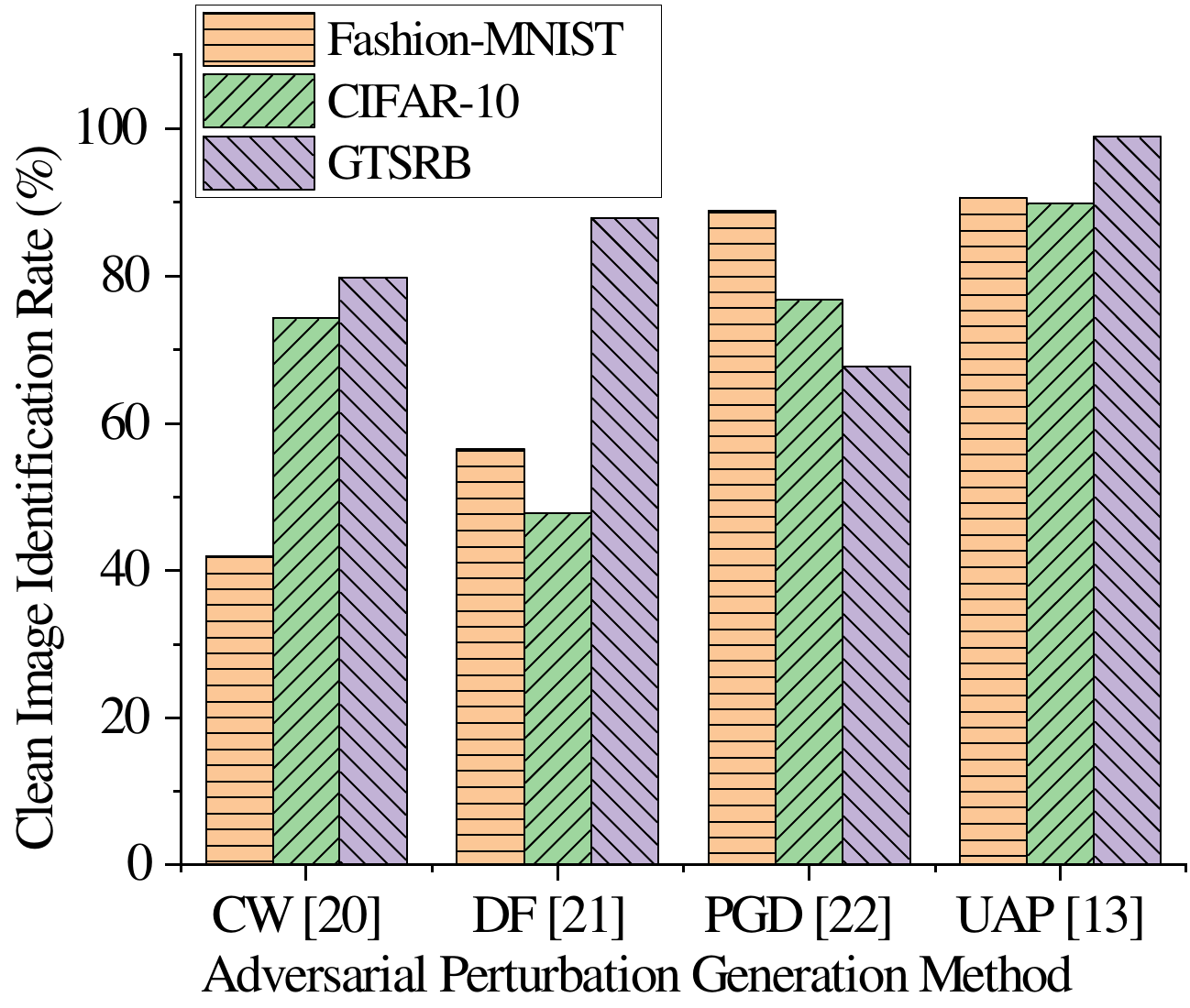}
\caption{The clean image identification rates of the proposed method by using different adversarial perturbation generation methods.}
\label{fig:ae2}
\end{figure}

In this section, the effectiveness of different adversarial perturbation generation methods is evaluated.
The four different adversarial perturbation generation methods evaluated in the experiment are C\&W \cite{Carlini017}, DeepFool \cite{Moosavi-Dezfooli16}, PGD \cite{MadryMSTV18} and UAP \cite{Moosavi-Dezfooli17}.
These four adversarial perturbation generation methods are separately used to generate the perturbation which is later utilized in the proposed method.
As shown in Fig. \ref{fig:ae1} and Fig. \ref{fig:ae2}, for the three datsets, using UAP \cite{Moosavi-Dezfooli17} in the proposed method obtains the highest BDR (99.63\%, 99.76\% and 99.91\% in Fashion-MNIST \cite{abs-1708-07747}, CIFAR-10 \cite{krizhevsky2009learning} and GTSRB \cite{StallkampSSI11} datasets, respectively), and the highest CIIR (90.66\%, 89.82\% and 98.85\% in Fashion-MNIST \cite{abs-1708-07747}, CIFAR-10 \cite{krizhevsky2009learning} and GTSRB \cite{StallkampSSI11} datasets, respectively) among all the four adversarial perturbation generation methods.
With the adversarial perturbation generated by C\&W \cite{Carlini017}, DeepFool \cite{Moosavi-Dezfooli16} and PGD\cite{MadryMSTV18}, the performance of the proposed method is less effective, as the backdoor detection rates are as low as 74.42\%, 65.22\% and 91.52\% by using C\&W \cite{Carlini017}, DeepFool \cite{Moosavi-Dezfooli16} and PGD \cite{MadryMSTV18}, respectively (as shown in Fig. \ref{fig:ae1}), and the clean image identification rates are as low as 41.92\%, 47.85\% and 67.77\% by using C\&W \cite{Carlini017}, DeepFool \cite{Moosavi-Dezfooli16} and PGD \cite{MadryMSTV18}, respectively (as shown in Fig. \ref{fig:ae2}).

As mentioned in Section \ref{why}, in the perturbation generation process, image-specific adversarial attack requires the label of each specific image to generate corresponding image-specific perturbation.
If the image is a backdoor instance, its label will be the target label, then the image-specific perturbation will be generated based on the target label instead of its ground-truth label.
Hence, the generated image-specific perturbation will strongly affect the backdoor trigger.
Once the backdoor trigger is strongly affected, it cannot trigger the backdoor, leading to the inconsistence of predicted labels of backdoor instance before and after perturbation.
As a result, this backdoor instance will be incorrectly considered as a benign image.
Therefore, the performance of the proposed method with image-specific perturbation is low.
UAP \cite{Moosavi-Dezfooli17} is a kind of image-agnostic perturbation, which is generated based on a small set of clean images.
The influence of UAP is mostly on the salient regions of the backdoor instance instead of the trigger.
Therefore, the predicted labels of the backdoor instance before and after perturbation are consistent, which makes the image be correctly detected as a backdoor instance.
In summary, among these four adversarial perturbation generation methods, UAP \cite{Moosavi-Dezfooli17} is most suitable for use in the proposed method.

\subsection{Comparison with Related Work}
\label{strip}
In this section, the proposed method is compared with STRIP \cite{GaoXW0RN19}.
In the detection process of STRIP \cite{GaoXW0RN19}, a set of other images from different classes are added to the input image separately in order to generate a set of blended images \cite{GaoXW0RN19}.
Then, STRIP utilizes entropy to measure the randomness of the predicted labels of all the blended images \cite{GaoXW0RN19}.
The entropy of clean images is significantly lower than the entropy of backdoor instances.
As a result, the smaller the entropy, the input image is more likely to contain a trigger \cite{GaoXW0RN19}.

The advantages of the proposed method over STRIP \cite{GaoXW0RN19} are as follows.
(i) The proposed method perturbs the image with universal perturbation \cite{Moosavi-Dezfooli17} rather than other images from different classes.
Since the $\ell_\infty$ norm of UAP \cite{Moosavi-Dezfooli17} is very low, the perturbation is very small.
In addition, because UAP is generated from clean images, the generated UAP mainly focuses on perturbing the salient regions of an image rather than the trigger.
Therefore, the trigger is almost unaffected.
However, in STRIP, other images are directly added to the input image, so the input image is globally affected \cite{GaoXW0RN19}.
It will not only destroy the main content of the input image, but also may break the trigger.
(ii) For each image, the proposed method only needs to generate one extra perturbed image and predict two images (perturbed and unperturbed image).
However, for each input image, STRIP needs to generate a set of blended images and input them to the model in order to estimate the entropy of the predicted labels of these blended images \cite{GaoXW0RN19}.
Therefore, the detection process of STRIP \cite{GaoXW0RN19} is more complex than the proposed method, thus the proposed method is more efficient than STRIP.

The experiment is also conducted to compare the proposed method with STRIP \cite{GaoXW0RN19}, and the experimental results are shown in Table \ref{tab:table5}.
We reproduce STRIP by following the method proposed in \cite{GaoXW0RN19} for comparision.
As shown in Table \ref{tab:table5}, the performance of the proposed method is significantly better than that of STRIP \cite{GaoXW0RN19} on all the three datasets.
Specifically, for STRIP, the backdoor detection rate is 63.40\%, 96.32\% and 73.95\% on Fashion-MNIST \cite{abs-1708-07747}, CIFAR-10 \cite{krizhevsky2009learning} and GTSRB \cite{StallkampSSI11} respectively.
For the proposed approach, the backdoor detection rate is 99.63\%, 99.76\% and 99.91\% on Fashion-MNIST \cite{abs-1708-07747}, CIFAR-10 \cite{krizhevsky2009learning} and GTSRB \cite{StallkampSSI11} datasets, respectively.
For the clean image identification rate, the proposed method also has better performance than STRIP \cite{GaoXW0RN19}.
Specifically, for STRIP, the clean image identification rate is 67.40\%, 89.57\% and 88.80\% on Fashion-MNIST \cite{abs-1708-07747}, CIFAR-10 \cite{krizhevsky2009learning} and GTSRB \cite{StallkampSSI11} datasets, respectively.
For the proposed method, the clean image identification rate is 90.66\%, 89.82\% and 98.85\% on Fashion-MNIST \cite{abs-1708-07747}, CIFAR-10 \cite{krizhevsky2009learning} and GTSRB \cite{StallkampSSI11} datasets, respectively.

\begin{table}[htbp]
  \centering
  \caption{Performance comparison between the proposed method and STRIP \cite{GaoXW0RN19}}
    \resizebox{\columnwidth}{!}{
    \begin{tabular}{ccccc}
    \toprule
    \multicolumn{1}{c}{\multirow{2}[4]{*}{\tabincell{c}{Dataset}}} & \multicolumn{2}{c}{\tabincell{c}{CIIR on \\ clean images}} & \multicolumn{2}{c}{\tabincell{c}{BDR on \\ backdoor instances}} \\
\cmidrule{2-5}          & STRIP \cite{GaoXW0RN19} & Ours  & STRIP \cite{GaoXW0RN19} & Ours \\
    \midrule
    \tabincell{c}{Fashion-MNIST} & 67.40\% & 90.66\% & 63.40\% & 99.63\% \\
    CIFAR-10 & 89.57\% & 89.82\% & 96.32\% & 99.76\% \\
    GTSRB & 88.80\% & 98.85\% & 73.95\% & 99.91\% \\
    \bottomrule
    \end{tabular}}%
  \label{tab:table5}%
\end{table}%

Note that, the intensity of trigger in this experiment is at a low level (0.15, 0.5, 0.2 for Fashion-MNIST \cite{abs-1708-07747}, CIFAR-10 \cite{krizhevsky2009learning} and GTSRB \cite{StallkampSSI11} datasets, respectively).
For STRIP \cite{GaoXW0RN19}, the backdoor instance is totally superimposed with other image from different classes, so the trigger is inevitably blended with other pixels.
Generally, when the intensity of trigger is normal, the trigger in the blended image may still activate the backdoor.
However, the intensity of trigger in this experiment is low, so the trigger in the blended backdoor instance is destroyed and will be ignored by the model.
Therefore, the entropy of this backdoor instance is similar to the entropy of clean images.
As a result, STRIP \cite{GaoXW0RN19} will incorrectly consider the backdoor instance as a clean one.
In comparison, the proposed method uses the universal adversarial perturbation (UAP \cite{Moosavi-Dezfooli17}) to perturb the input image.
First, unlike STRIP \cite{GaoXW0RN19}, adversarial perturbation will not globally perturb the input image.
The perturbation will only modify limited number of pixels as the $\ell_\infty$ norm of UAP is very low.
Second, UAP \cite{Moosavi-Dezfooli17} is generated from a small set of clean images.
Therefore, even if the backdoor instance is perturbed, the trigger in the backdoor instance will only be slightly affected.
As a result, the proposed method can successfully detect backdoor instances carrying the trigger with low intensity.
However, STRIP \cite{GaoXW0RN19} fails to detect some backdoor instances in this experiment, where the intensity of trigger is at a low level.

In summary, there are two advantages of the proposed method over STRIP \cite{GaoXW0RN19}.
First, the the proposed method is more effective than STRIP, as the proposed method will not destroy the trigger while STRIP may destroy the trigger.
Second, the proposed method is more efficient than STRIP, as the proposed method only needs to predict two images (perturbed and unperturbed image) for each untrusted image and STRIP needs to predicts a set of blended images.

\section{Conclusion}
\label{sec:Conclusion}
In this paper, we propose a novel backdoor detection method based on adversarial perturbations.
Specifically, the universal adversarial perturbation \cite{Moosavi-Dezfooli17} is first generated from the model, then the generated perturbation is added to the image.
If the prediction of model on the perturbed image is consistent with the one on the unperturbed image, the input image is considered as a backdoor instance.
Experimental results show that, the proposed defense method can achieve high backdoor detection rate and high clean image identification rate, while maintaining the visual quality of the image.
Besides, the defense performance of the proposed method against backdoor attacks under different settings is also demonstrated to be effective.
Our future work will explore the defenses against physical backdoor attacks in real physical world.

\ifCLASSOPTIONcaptionsoff
  \newpage
\fi

\bibliographystyle{IEEEtran}
\bibliography{ref}

\end{document}